\documentclass[english]{llncs}
\usepackage{mathtools}
\usepackage{graphicx}
\usepackage{caption}
\usepackage{subcaption}
\usepackage{gensymb}
\usepackage{amsmath}
\usepackage{babel}
\usepackage{algorithm}
\usepackage{algpseudocode}
\usepackage{algorithmicx}
\usepackage{siunitx}
\usepackage{booktabs} 
\usepackage{epstopdf}
\usepackage{setspace}
\usepackage{threeparttable}
\usepackage{booktabs}

\usepackage{subcaption}
\usepackage{etoolbox}
\usepackage{lipsum}
\let\oldbibliography\bibliography
\renewcommand{\bibliography}[1]{{%
  \let\chapter\section
  \oldbibliography{#1}}}

\makeindex 
\begin{document}
\title{Automatic segmentation of the intracranial volume  in fetal MR images }
\author{N. Khalili$^1$, P. Moeskops$^2$, N.H.P. Claessens$^3$, S. Scherpenzeel$^3$, E. Turk$^3$, R. de Heus$^4$, M.J.N.L. Benders$^{3,5}$, M.A. Viergever$^{1,5}$, J.P.W. Pluim$^{1,2}$, I. I\v sgum$^{1,5}$ }
 
\institute{Image Sciences Institute, University Medical Center Utrecht, Utrecht, The Netherlands
\and Medical Image Analysis Group, Department of Biomedical Engineering, Eindhoven University of Technology, Eindhoven, The Netherlands\\
\and Department of Neonatology, Wilhelmina Children’s Hospital, University Medical Center Utrecht, Utrecht, The Netherlands\\
\and Department of Obstetrics, University Medical Center Utrecht, Utrecht, The Netherlands\\
\and Brain Center Rudolf Magnus, University Medical Center Utrecht, Utrecht, The Netherlands}
\maketitle
\begin{abstract}
 
MR images of the fetus allow non-invasive analysis of the fetal brain. Quantitative analysis of fetal brain development requires automatic brain tissue segmentation that is typically preceded by segmentation of the intracranial volume (ICV). This is challenging because fetal MR images visualize the whole moving fetus and in addition partially visualize the maternal body. 
This paper presents an automatic method for segmentation of the ICV in fetal MR images. The method employs a multi-scale convolutional neural network in 2D slices to enable learning spatial information from larger context as well as detailed local information. 
The method is developed and evaluated with 30 fetal T2-weighted MRI scans (average age $33.2\pm1.2$ weeks postmenstrual age). The set contains $10$ scans acquired in axial, $10$ in coronal and $10$ in sagittal imaging planes. A reference standard was defined in all images by manual annotation of the intracranial volume in $10$ equidistantly distributed slices.
The automatic analysis was performed by training and testing the network using scans acquired in the representative imaging plane as well as combining the training data from all imaging planes. On average, the automatic method achieved Dice coefficients of 0.90 for the axial images, 0.90 for the coronal images and 0.92 for the sagittal images.  Combining the training sets resulted in average Dice coefficients of 0.91 for the axial images, 0.95 for the coronal images, and 0.92 for the sagittal images.
The results demonstrate that the evaluated method achieved good performance in extracting ICV in fetal MR scans regardless of the imaging plane.

\end{abstract}

\section{Introduction}Fetal magnetic resonance imaging (MRI) is increasingly used as a non-invasive tool for monitoring the fetal brain development. A number of papers describing automatic quantitative analysis of brain in MR scans of fetuses have been published \cite{gholipour2011fetal}\cite{gholipour2017normative}\cite{wright2014automatic}. Compared to automatic analysis of the brain in adults and neonates, analysis of the fetal brain carries unique challenges. In contrast to brain MRI of adults and neonates that mostly visualize the head only, fetal MR images have a larger field of view that includes the entire fetus as well as part of the maternal body. In addition, because of fetal movement, the brain location and orientation in fetal MRI may vary considerably. Hence, prior to analysis of the brain tissue classes, automatic methods often determine a volume of interest containing the brain.
 
To identify a volume of interest, several methods performed brain segmentation.
Anquez et al. \cite{anquez2009automatic} proposed a method for automatic segmentation of intracranial volume (ICV) in fetal MR scans. The method first localizes the eyes and exploits this information to segment ICV using a graph cut approach that is guided by shape, contrast and biometrical priors. The method was applied to scans with unknown fetal orientation and the results demonstrate that it is able to perform segmentation with high accuracy.  
Rajchl et al. \cite{rajchl2017deepcut} proposed a deep learning approach for brain and lung segmentation. Their method combines a convolutional neural network and iterative graphical optimization to obtain the final segmentation. The method is trained with weakly labeled data consisting of the brain bounding boxes. It was applied to data with large anatomical variation and achieved high segmentation accuracy. 
Recently, Salehi et al. \cite{salehi2017auto} proposed a method for segmentation of the brain in adult and fetal MR scans. In this approach, the fetal brain is first localized by defining a bounding box around it with ITKSNAP \cite{yushkevich2006user}. Next, the segmentation was performed using a multi-scale CNN as proposed by Moeskops et al. \cite{moeskops2015automatic} with an iterative approach that uses input from the posterior probabilities of the previous segmentation step to refine the segmentations. The authors reported high segmentation accuracy.
 
Unlike methods performing ICV segmentation, several methods perform brain localization in fetal MRI.
Ison et al.\cite{ison2012fully} proposed a pipeline to detect a bounding box in several stages. The method first employs a two-stage random forest classifier. In the first stage, maternal tissue is separated from the fetal head, and in the second stage the fetal brain tissue is classified in several classes. Thereafter, a Markov random field appearance model is used to find the brain orientation using results of the brain tissue classification. Hence, the detected bounding box follows the orientation of the brain in the scan. The results show that the proposed approach is robust but with moderate accuracy. Only 28\% of the coronal images and 53\% of the axial and sagittal images contained whole brain in the detected bounding box.
Keraudren et al. \cite{keraudren2013localisation} proposed a method for brain localization that determines a bounding box in the orientation of the image axis. The method first fits an ellipse around the brain in every image slice. Thereafter, ellipses meeting criteria about expected brain size and knowledge about gestational age are analyzed using SIFT features in a bag-of-words model to identify brain voxels. Thereafter, a bounding box is fitted to the extracted feature cloud using a RANSAC algorithm. The method can be applied to scans of any orientation and it achieved good results with $85\%$ of the cases containing the whole brain within the bounding box. 
Taimouri et al. \cite{taimouri2015template} proposed a template matching approach to find slices containing the brain. This was applied to determine a bounding box around the brain. Unlike methods that encode prior knowledge about the gestational age and brain size in the features, this method uses prior information from an age-matched template. The results demonstrate high success rate in determination of the brain bounding boxes.

 In this study, we investigated whether the method previously developed for brain tissue segmentation by Moeskops et al. \cite{moeskops2015automatic} can be used for the challenging task of segmentation of the ICV in fetal MRI. Unlike other methods for fetal brain segmentation, the proposed method directly segments the ICV from MRI data without the need to crop the image to a region of interest first or using prior knowledge about the brain size or gestational age. We evaluated the method with MR scans acquired in axial, coronal and sagittal imaging planes.

\section{Data}
 
This study includes T2-weighted MR scans of 10 fetuses. The gestational age of fetuses ranged from $22.9$ to $34.6$ weeks. For every patient, 3 images were acquired on a Philips Achieva 3T scanner using a turbo fast spin-echo sequence. The data set contains 30 images in total: 10 images acquired in the axial, 10 images in the coronal and 10 images in the sagittal imaging plane. The acquired voxel size is $1.25\times1.25\times2.5$ mm$^3$ and reconstructed voxel size is $0.7\times0.7\times2.5$ mm$^3$. The reconstruction matrix is $512\times512\times80$.

To define the reference standard, manual segmentation of the ICV was performed by a trained medical student in 10 slices for each of the 30 MR images. The slices were equidistantly distributed over the brain. Manual annotation was performed using in-house developed software. The process was done by painting brain voxels in each slice.

Images were divided into training and test set. The training set contained images of 7 patients and the test set contained images of the remaining 3 patients.
  
	\begin{figure}[t]	
		\begin{centering}
			\includegraphics[width=1.0\textwidth]{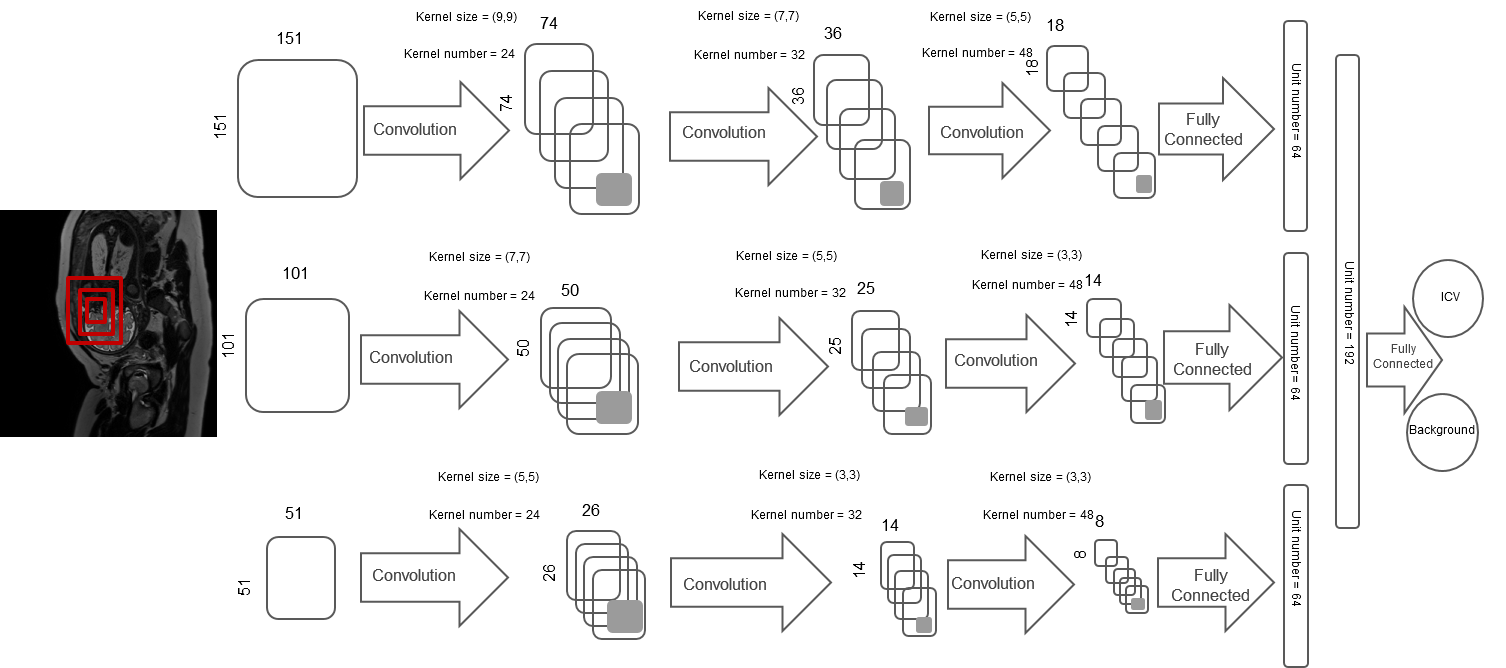}
			\par\end{centering}
		\caption{\label{fig:CNN} CNN architecture: The network contains three branches which are fed by three different patch sizes extracted from the input image as illustrated by red squares in the input image. Each branch has three convolution layers and two fully connected layers. In the last layer, all three branches are concatenated in a single fully connected layer and classified as either ICV or background.}
	\end{figure}

\section{Method}

To segment ICV a multi-scale convolutional neural network (CNN) as proposed by Moeskops et al. \cite{moeskops2015automatic} is employed. This network has been developed for brain tissue segmentation in neonatal and adult brain MR scans. The network analyzes 2D patches of different sizes to allow exploiting information from the local and global context. Hence, the network takes three patch sizes that are each analyzed in a separate network branch and combined in the last layer. Each branch consists of three convolutional layers alternated with downsampling layers to reduce feature map sizes. Convolution layers of each network branch are followed by a fully connected layer. Outputs of the three branches are concatenated and input for a fully connected layer with a softmax function, which distinguishes between positive (ICV) and negative (mother and fetus excluding the brain) classes. As opposed to the network used for neonatal and adult images \cite{moeskops2015automatic}, the network in the current study uses larger input patches ($51\times51$, $101\times101$ and $151\times151$ vs. $25\times25$, $51\times51$ and $75\times75$), and uses strided convolution instead of maxpooling. Detailed network parameters are shown in Figure \ref{fig:CNN}. 
To avoid bias towards the majority class, the network was trained with an equal number of positive (ICV) and negative (mother and fetus excluding the brain) samples randomly selected from each scan. In addition, training samples were randomly chosen in every epoch. In order to avoid overfitting, batch normalization \cite{ioffe2015batch} was used for the convolutional layers  and dropout \cite{srivastava2014dropout} was used for the fully connected layers. The CNN was trained by backpropagation using cross-entropy as loss function and Adam \cite{kingma2015method} for optimizing the weights and biases. 
 
Pixel classification may result in isolated clusters of voxels that may locally resemble the brain. To prevent this, only the largest 3D connected component segmented by the CNN was retained.

\section{Experiments and Results}
Three sets of experiments were performed. In all experiments, segmentation performance was evaluated using Dice coefficients between manually and automatically segmented images. 
 
First, to evaluate the performance of the network when training and testing with the representative data, the network was trained with images acquired in axial, coronal and sagittal planes separately and tested with images from the corresponding orientation. From each set of axial, coronal and sagittal scans, 7 images were used for training, and the remaining 3 images from each orientation were used for testing. 
 
Second, to evaluate whether the network is able to generalize with respect to image orientation, the network was trained with a mix of images from the axial, coronal and sagittal planes, and evaluated with images acquired in all three orientations. Thus, training and test sets from the first set of experiments were merged. Hence, the training set contained 21 images and test set contained 9 images.
 
Third, because the second  experiment uses a much larger training set than the first experiment (21 vs. 7 training images), experiments with 7 training images acquired in axial, coronal and sagittal orientations were performed as well. For this purpose, experiments with three different training settings were performed: From a set of 3 axial, 3 coronal and 3 sagittal training scans, training was performed using 3 axial, 2 coronal and 2 sagittal images; 2 axial, 3 coronal and 2 sagittal images; and 2 axial, 2 coronal and 3 sagittal images. 

\begin{figure}
\begin{subfigure}{.33\textwidth}
  \centering
  \includegraphics[width=\linewidth,scale=1]{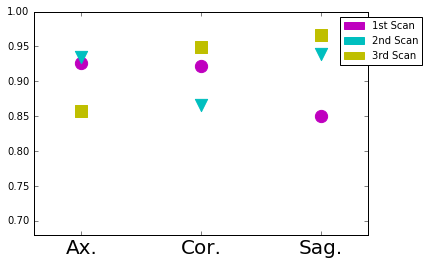}
  \caption{}
  \label{fig:sfig1}
\end{subfigure}%
\begin{subfigure}{.33\textwidth}
  \centering
  \includegraphics[width=\linewidth,scale=0.3]{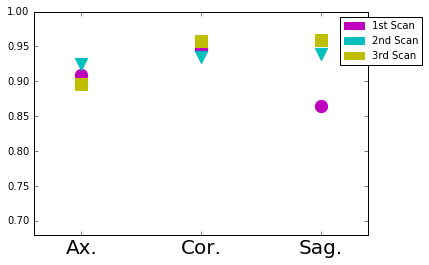}
  \caption{}
  \label{fig:sfig2}
 
\end{subfigure}
\begin{subfigure}{.33\textwidth}
  \centering
  \includegraphics[width=\linewidth,scale=0.3]{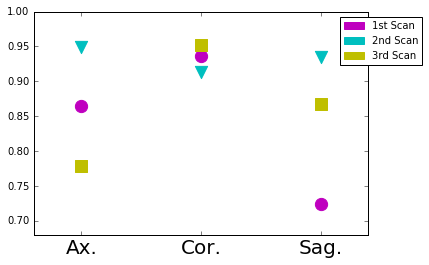}
  \caption{}
  \label{fig:sfig1}
\end{subfigure}%
\begin{subfigure}{.33\textwidth}
  \centering
  \includegraphics[width=\linewidth,scale=0.3]{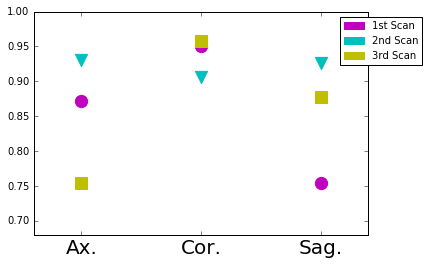}
  \caption{}
  \label{fig:sfig2}
\end{subfigure}
\begin{subfigure}{.33\textwidth}
  \centering
  \includegraphics[width=\linewidth,scale=0.3]{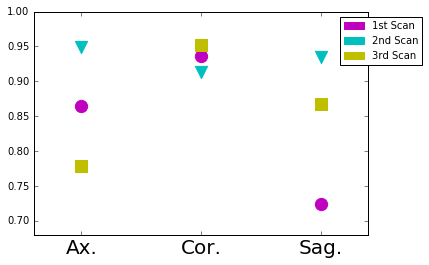}
  \caption{}
  \label{fig:sfig2}
\end{subfigure}
\caption{Dice coefficients obtained between manual and automatic segmentation in different experimental settings for each test scan in axial (Ax.), coronal (Cor.) and sagittal (Sag.) imaging plane. Results were obtained when the CNN was trained with: (a) 7 scans from a single orientation and tested with scans acquired in the representative orientation; (b) a combination of  7 axial, 7 coronal, 7 sagittal scans; (c) 3 axial, 2 coronal and 2 sagittal; (d)  2 axial, 3 coronal and 2 sagittal scans; (e) 2 axial, 2 coronal and 3 sagittal scans. }
\label{fig:plot}
 
\end{figure}

\begin{figure}
\begin{subfigure}{.33\textwidth}
  \centering
  \includegraphics[width=0.89\linewidth,height = 0.875\linewidth,scale=0.3]{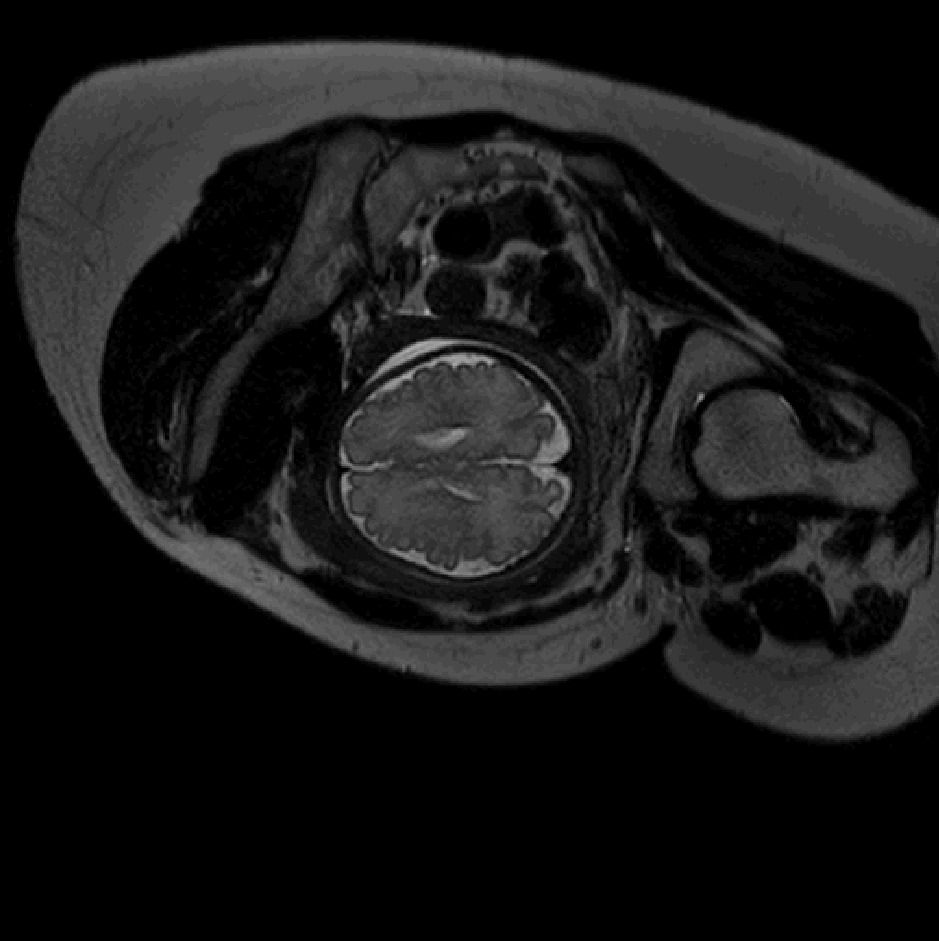}
  \label{fig:sfig1}
\end{subfigure}%
\begin{subfigure}{.33\textwidth}
  \centering
  \includegraphics[width=0.89\linewidth,height = 0.875\linewidth,scale=0.3]{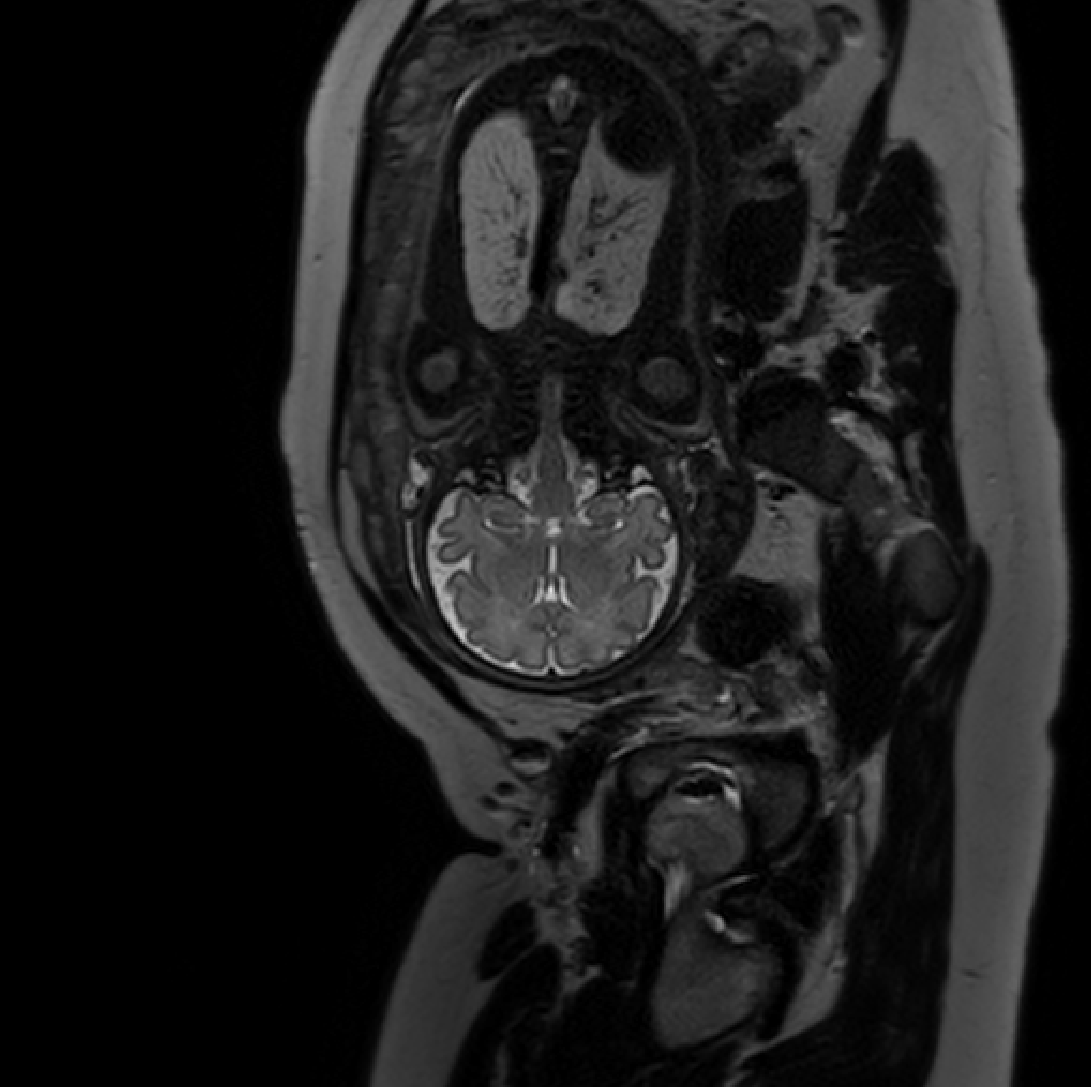}
  \label{fig:sfig2}
\end{subfigure}
\begin{subfigure}{.33\textwidth}
  \centering
  \includegraphics[width=0.88\linewidth,scale=0.3]{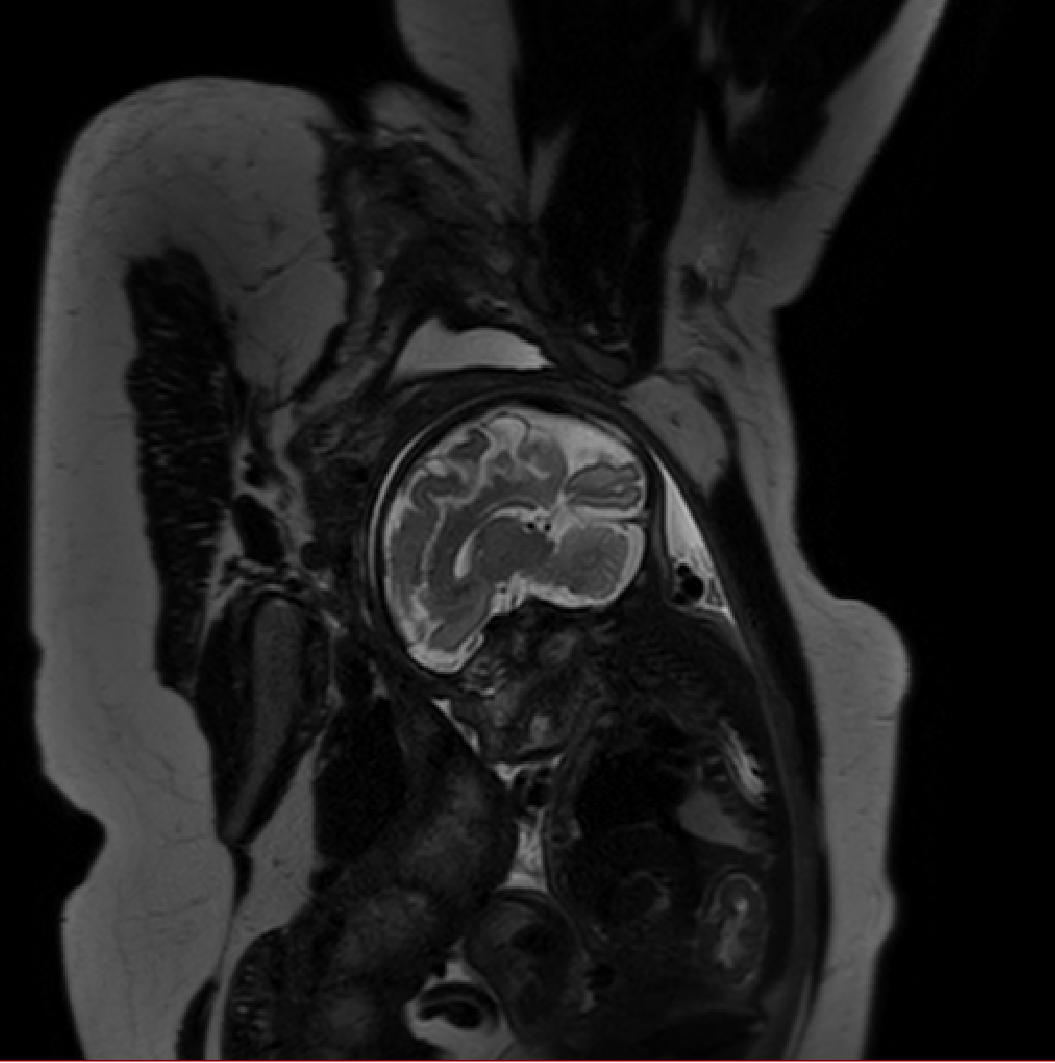}
  \label{fig:sfig2}
\end{subfigure}
\begin{subfigure}{.33\textwidth}
  \centering
  \includegraphics[width=0.9\linewidth,scale=0.3]{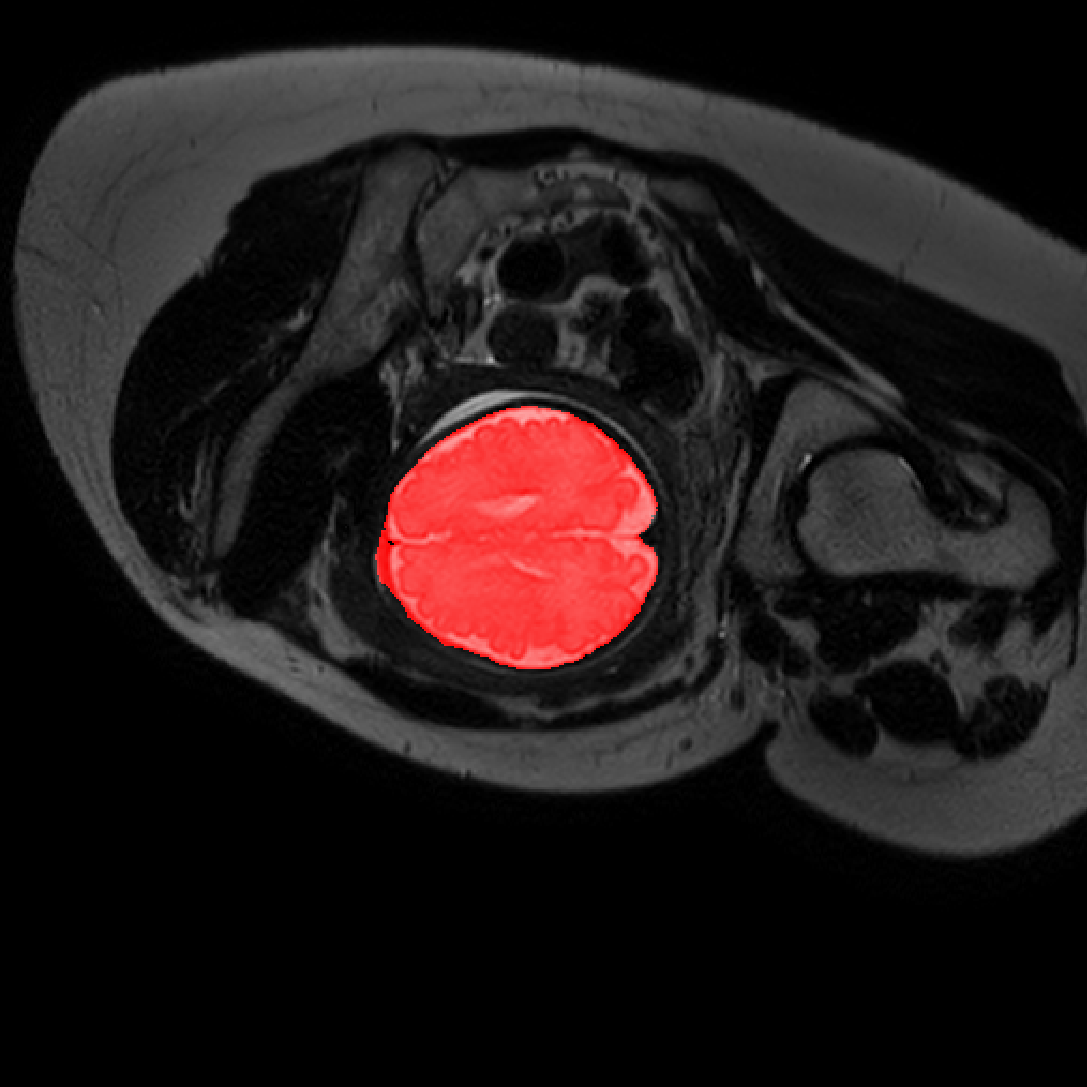}
  \label{fig:sfig1}
\end{subfigure}%
\begin{subfigure}{.33\textwidth}
  \centering
  \includegraphics[width=0.89\linewidth,scale=0.3]{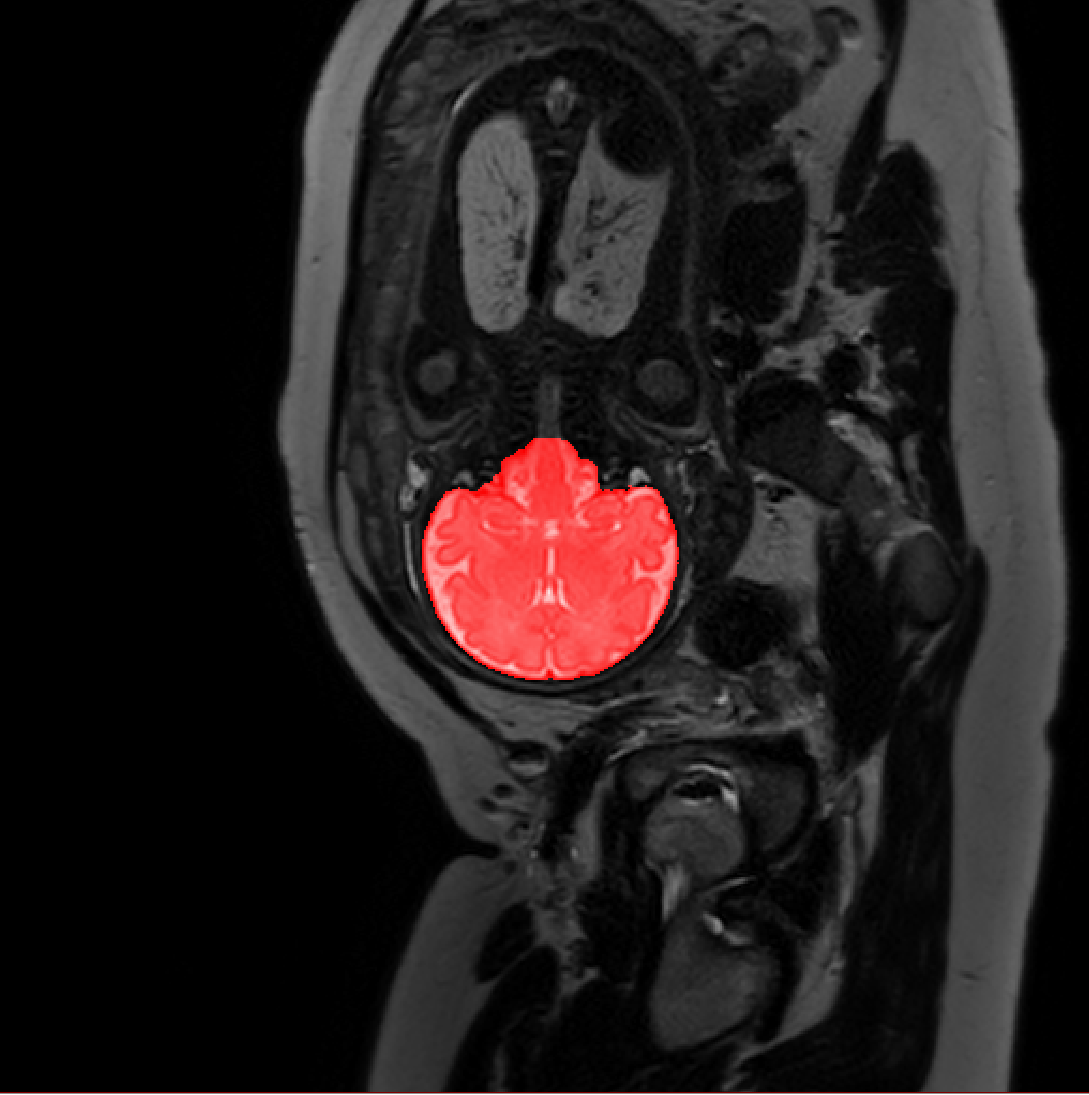}
  \label{fig:sfig2}
\end{subfigure}
\begin{subfigure}{.33\textwidth}
  \centering
  \includegraphics[width=0.89\linewidth,scale=0.3]{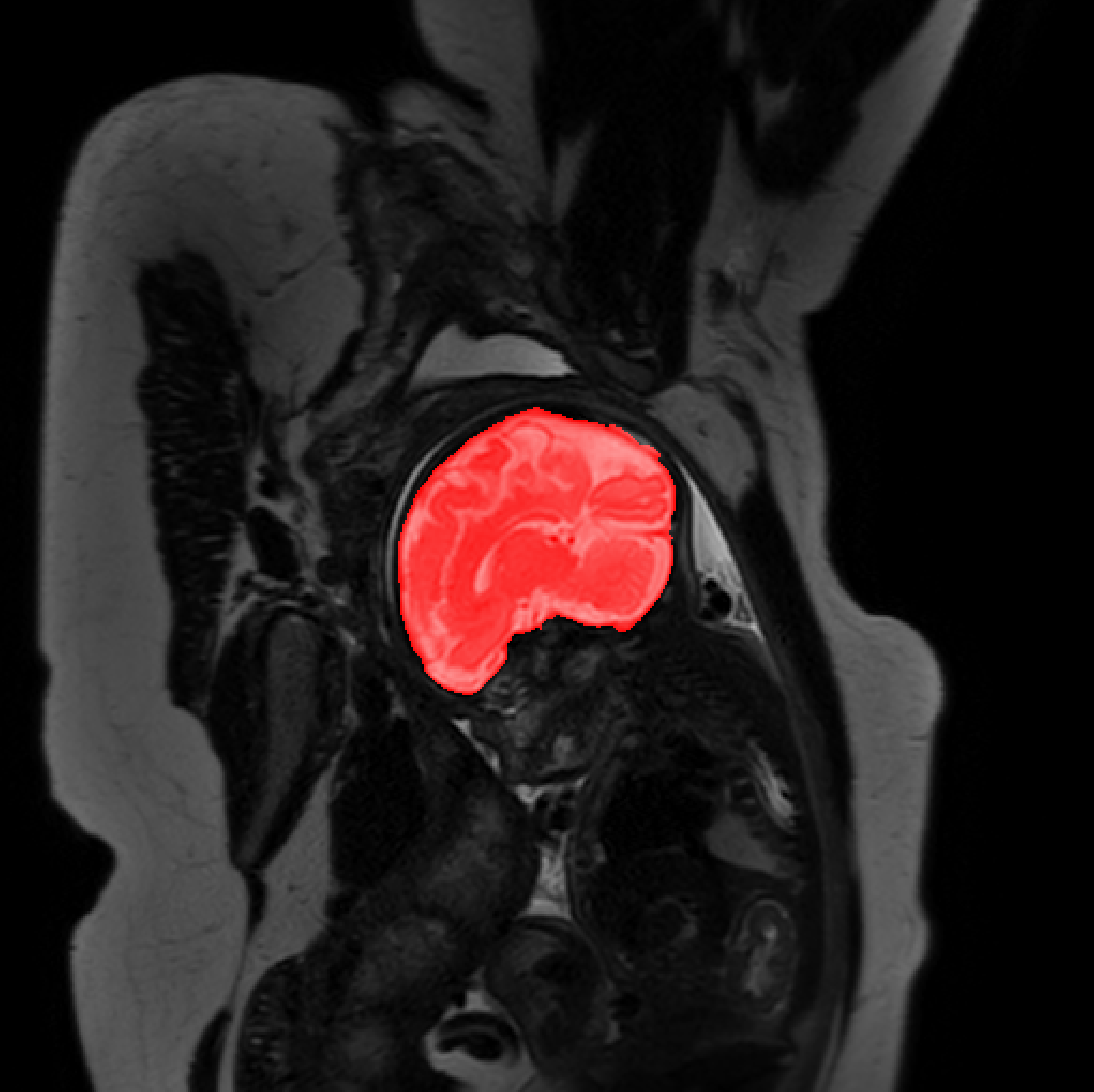}
  \label{fig:sfig2}
\end{subfigure}
\begin{subfigure}{.33\textwidth}
  \centering
  \includegraphics[width=0.9\linewidth,scale=0.3]{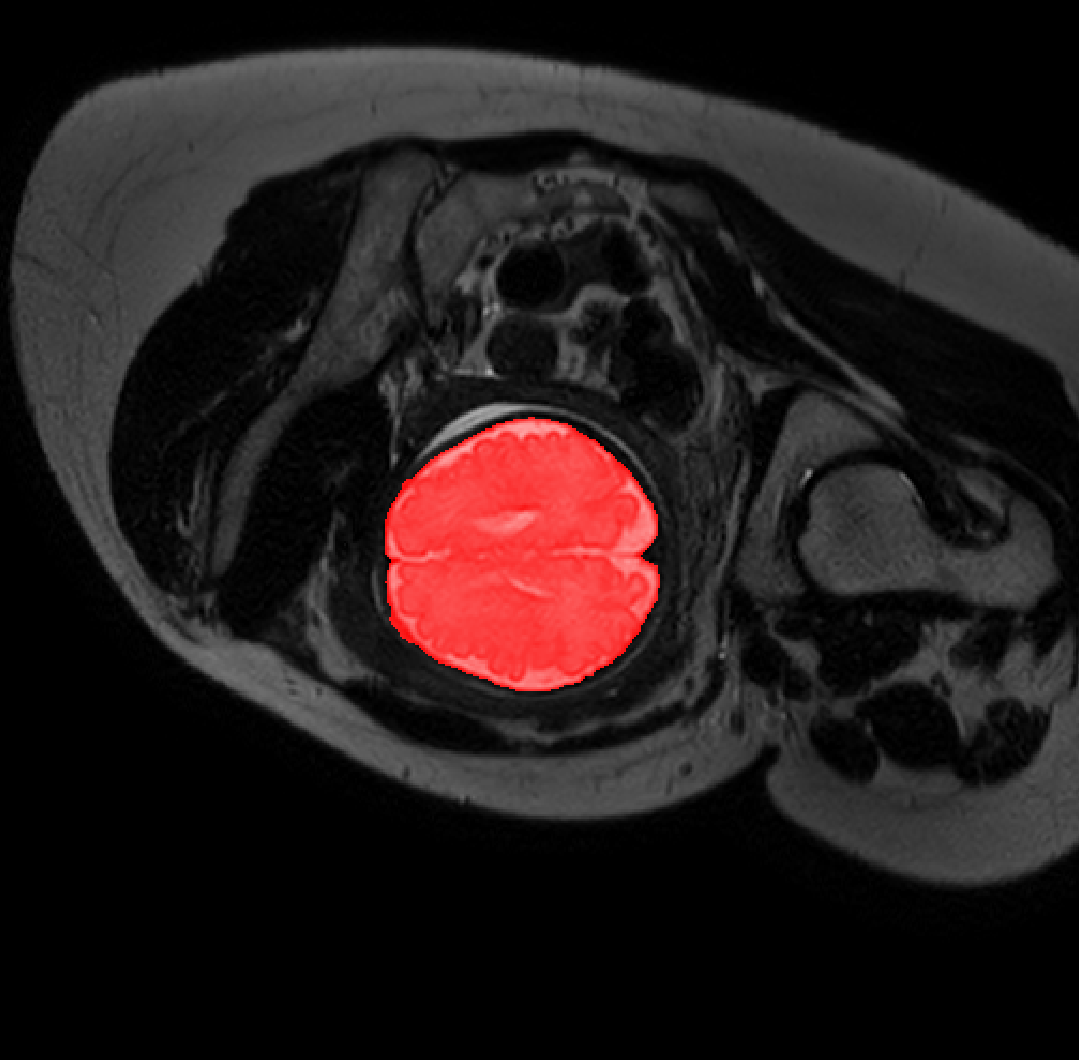}
  \label{fig:sfig1}
\end{subfigure}%
\begin{subfigure}{.33\textwidth}
  \centering
  \includegraphics[width=0.89\linewidth,scale=0.3]{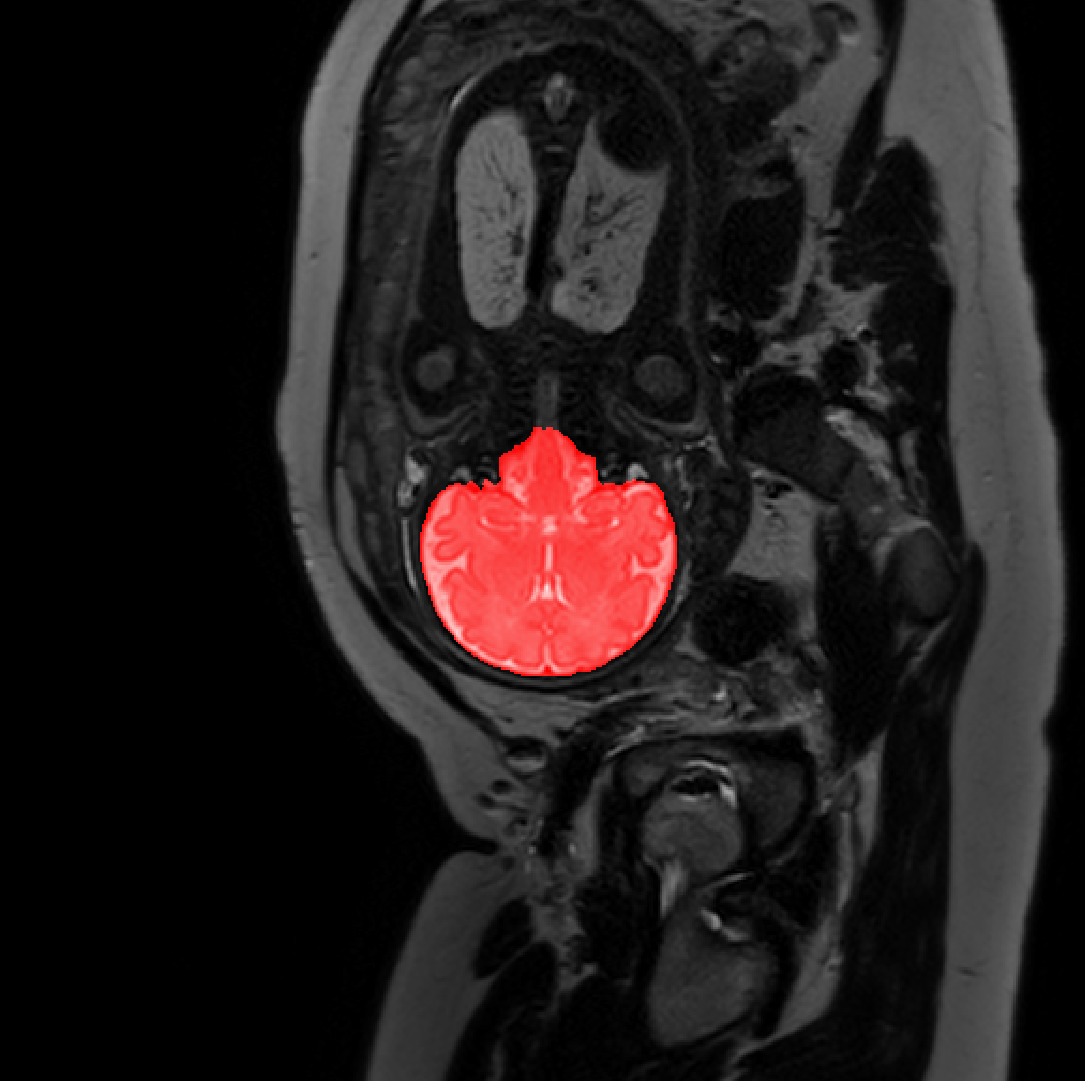}
  \label{fig:sfig2}
\end{subfigure}
\begin{subfigure}{.33\textwidth}
  \centering
  \includegraphics[width=0.89\linewidth,scale=0.3]{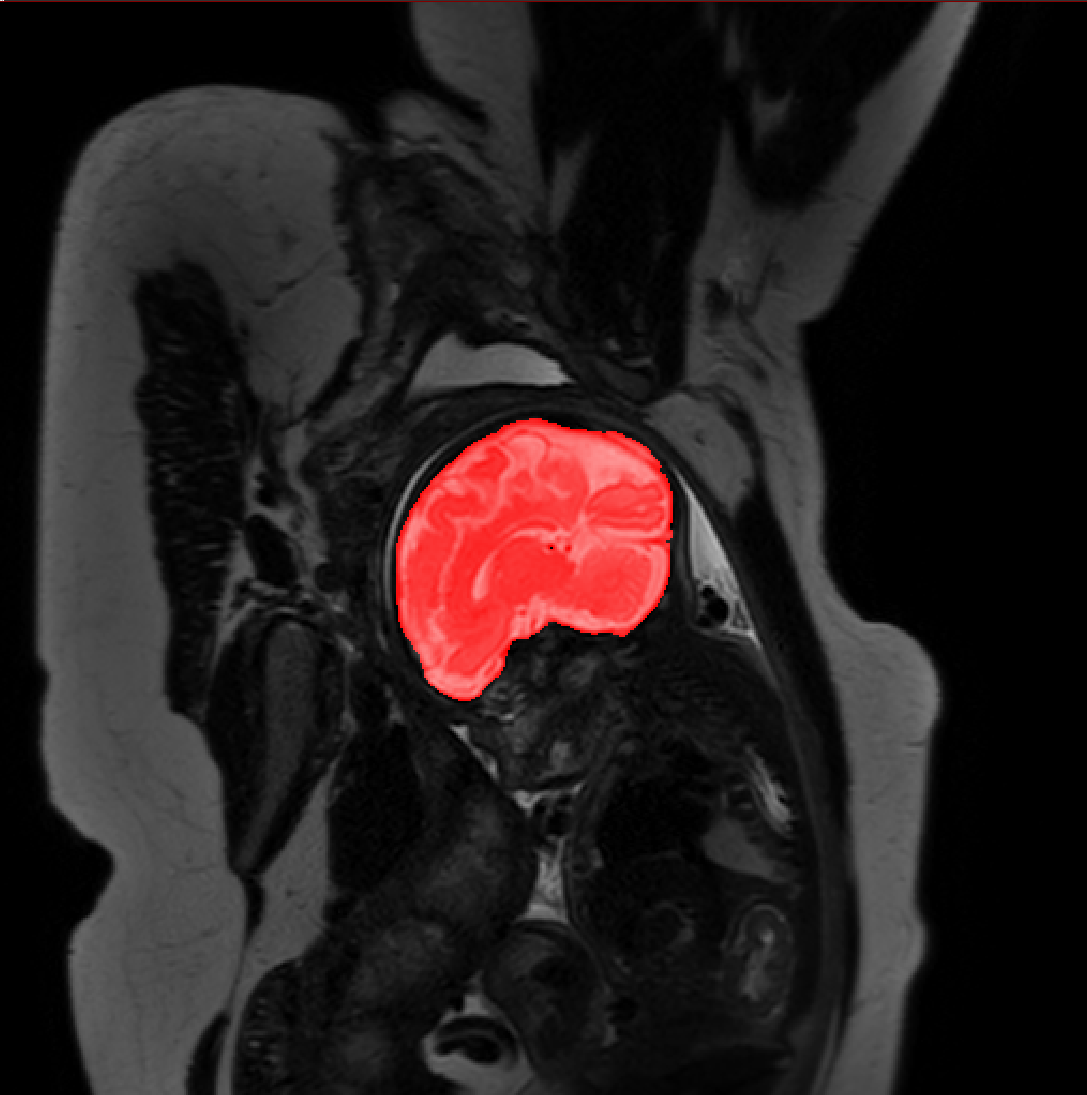}
  \label{fig:sfig2}
\end{subfigure}
\begin{subfigure}{.33\textwidth}
  \centering
  \includegraphics[width=0.9\linewidth,scale=0.3]{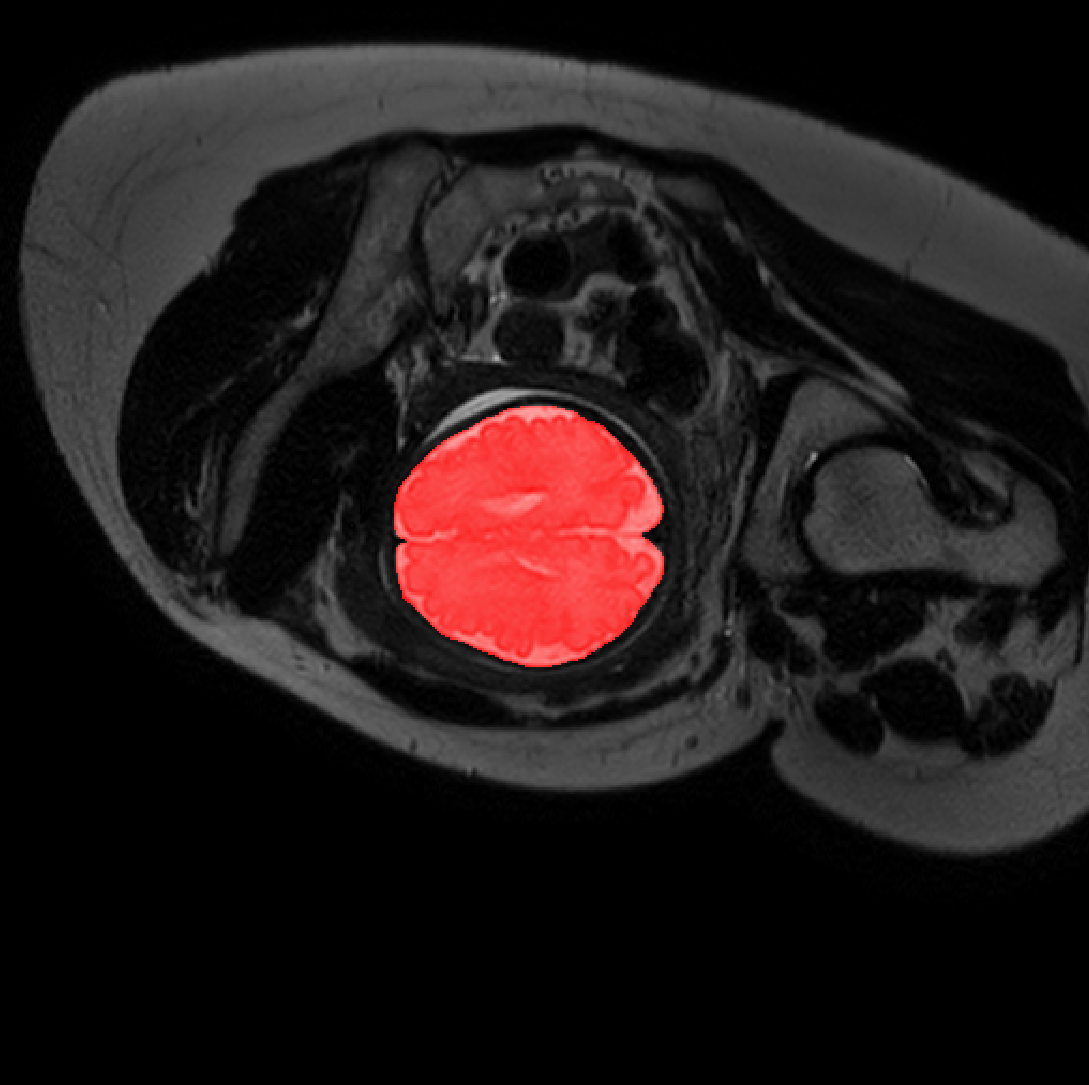}
  \label{fig:sfig1}
\end{subfigure}%
\begin{subfigure}{.33\textwidth}
  \centering
  \includegraphics[width=0.89\linewidth,scale=0.3]{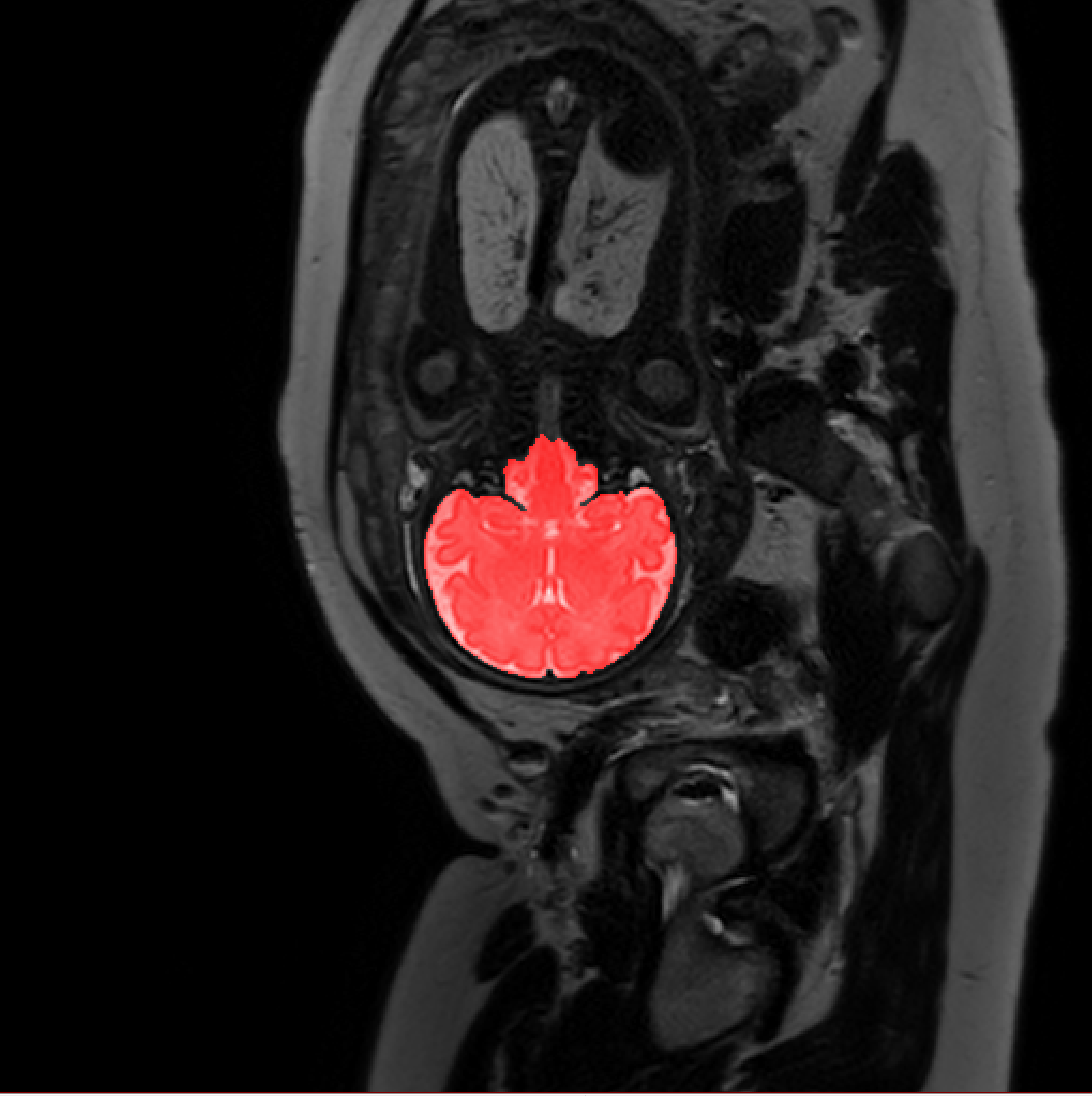}
  \label{fig:sfig2}
\end{subfigure}
\begin{subfigure}{.33\textwidth}
  \centering
  \includegraphics[width=0.89\linewidth,scale=0.3]{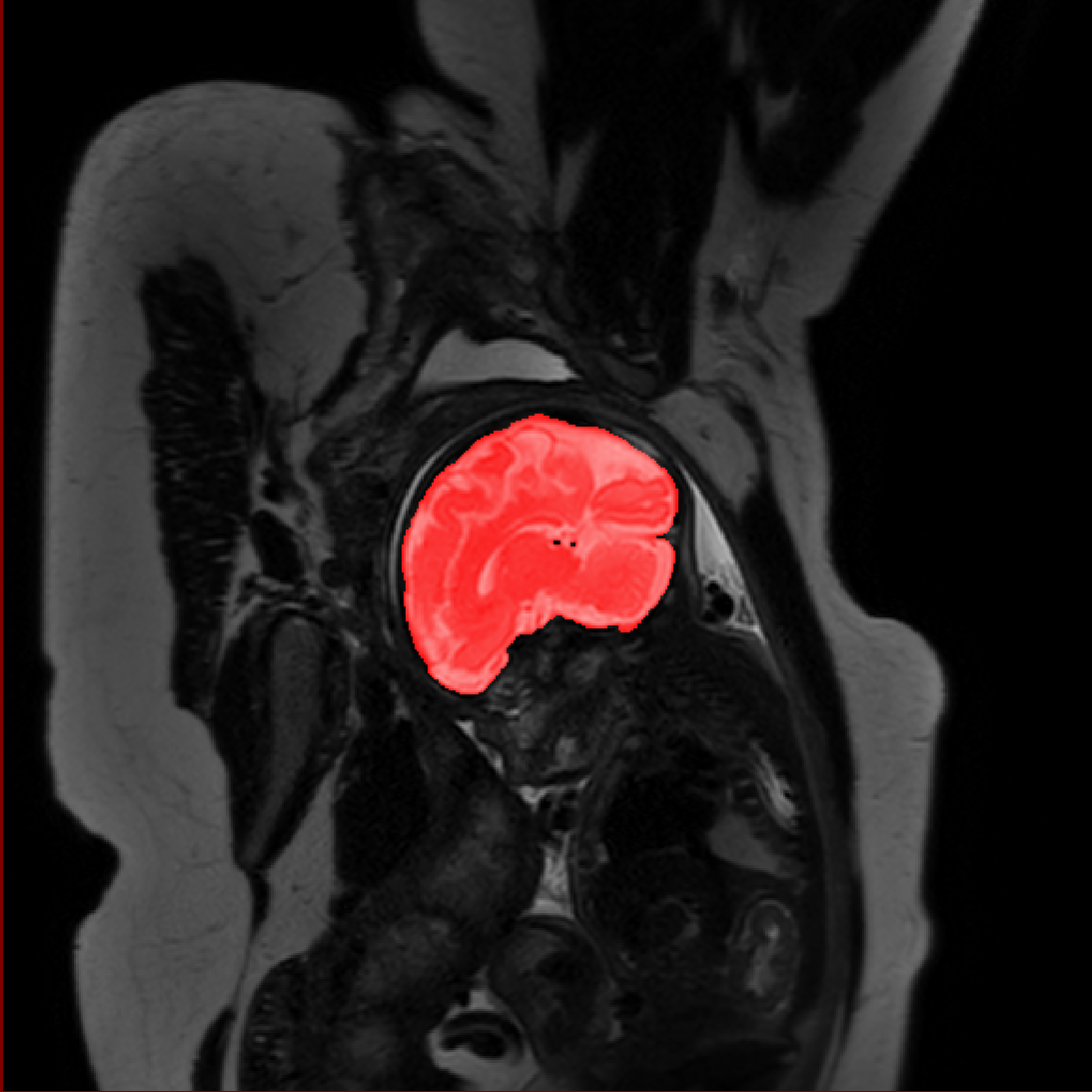}
  \label{fig:sfig2}
\end{subfigure}
 
\caption{Example of ICV segmentations in images acquired in axial (left), coronal (middle) and sagittal (right)
planes. Top row: A slice from T2-weigted image; Second row: Automatic segmentations obtained using 7 training images from the representative imaging planes; Third row: Automatic segmentations obtained using all 21 training images from all 3 image orientations; Bottom row: Manual segmentation. }
\label{fig:qualitative}
\end{figure}

Table~\ref{tab:dice} lists average of quantitative evaluation results of these experiments and Figure~\ref{fig:plot} shows results obtained from each image. Figure \ref{fig:qualitative} shows examples of the obtained segmentations.

\begin{table}[htp]

\centering
\footnotesize\setlength{\tabcolsep}{2.5pt}
\begin{tabular}{l@{\hspace{6pt}} *{3}{c}}
\toprule
 \multicolumn{2}{c}{} &
 \multicolumn{1}{c}{Test set}\\
Training set composition

& Axial& Coronal  & Sagittal  \\
\midrule
 7 axial
&0.90&-&- \\
 7 coronal
&-&0.90&- \\
 7 sagittal
&-&-&0.92 \\
 7 axial, 7 coronal, 7 sagittal
& 0.91&
0.95
&0.92 \\
 
 3 axial, 2 coronal, 2 sagittal
& 0.86 & 0.93 & 0.84 \\
 
 2 axial, 3 coronal, 2 sagittal
&0.85 & 0.94 & 0.85 \\
 2 axial, 2 coronal, 3 sagittal
& 0.89 &0.93& 0.84 \\

\bottomrule
\addlinespace

\end{tabular}

\caption{Average Dice coefficients between manually and automatically obtained ICV segmentations:
First three rows  list results when network was trained with 7 images from a single plane and tested with images acquired in the corresponding plane. Fourth row lists results obtained when the network was trained with 21 training images from all three imaging planes. Last three rows list  results of the network trained with 7 training images combined from all three imaging planes. }
 
 \label{tab:dice}
\end{table}

\begin{figure}
\begin{subfigure}{.33\textwidth}
  \centering
  \includegraphics[width=0.85\linewidth,scale=0.2]{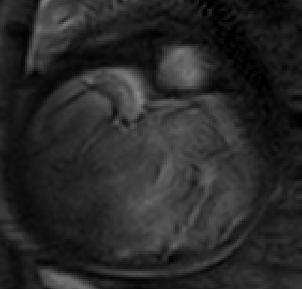}
  \label{fig:motion}
\end{subfigure}%
\begin{subfigure}{.33\textwidth}
  \centering
  \includegraphics[width=0.85\linewidth,scale=0.2]{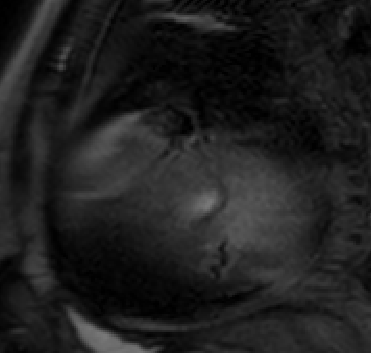}
  \label{fig:sfig2}
\end{subfigure}
\begin{subfigure}{.33\textwidth}
  \centering
  \includegraphics[width=0.82\linewidth,scale=0.2]{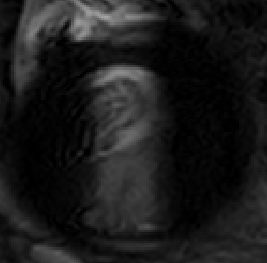}
  \label{fig:sfig2}
\end{subfigure}
\caption{Example of three slices strongly affected by imaging artifacts.}
\label{fig:motion_seg}
 
\end{figure}

 \begin{table}[htp]

\centering
\footnotesize\setlength{\tabcolsep}{2.5pt}
\begin{tabular}{l@{\hspace{6pt}} *{3}{c}}
\toprule
Previous studies 
& Dice Coefficients & Hausdorff Distance(mm)  & Kappa  \\
\midrule

 Salehi et al. \cite{salehi2017auto}
&0.98&-&- \\
 Rajchl et al. \cite{rajchl2017deepcut}
&0.94&-&- \\
 Anquez et al.\cite{anquez2009automatic}
& -&
3.4
&0.93 \\

\bottomrule
\addlinespace

\end{tabular}
 
\caption{ Segmentation performance as reported in three previous studies. Note that these results are obtained on different data sets, and can therefore not be directly compared.}
 
 \label{tab:compare}
\end{table}
 
\section{Discussion}
We have presented an evaluation of a multi-scale CNN for segmentation of ICV in fetal MRI. The method was evaluated with images acquired in axial, coronal and sagittal imaging planes and the results demonstrate that the method achieves Dice coefficients of $0.91$ regardless of the image orientation. Unlike previous fetal brain extraction methods, the proposed method segments ICV from fetal MRI without the need for bounding box localization or exploiting prior information about the patient age or expected anatomy. To allow an indication of the segmentation performance compared with other methods, the results from previous studies are summarized in Table \ref{tab:compare}. Note that these results are obtained on different data sets, and can therefore not be directly compared.
 
The segmentation performance appears similar when using representative training images and when using mixed training images with respect to imaging plane. This demonstrates that the method is robust with respect to image orientation. Results obtained in the experiment using a larger training set with training images from all three imaging planes do not seem to lead to substantial improvement in performance, although they tend to be more consistent. Because a small test set of in total 9 test scans from 3 patients was available, statistical difference among the different experimental settings was not evaluated. In future research these results need to be confirmed in a larger set of images.
 
Because of the continuous fetal motion, fetal MR images contain motion artifacts that were most present in scans acquired in sagittal imaging plane. Slices that contained very strong artifacts were not segmented by the automatic method (Figure \ref{fig:motion_seg}). Because quantitative evaluation was performed only in slices with manual annotations, these did not affect the quantitative evaluation. Nevertheless, poor image quality in such slices prohibits manual expert as well as automatic brain segmentation. Hence, to solve this, motion correction, as e.g. proposed by Kainz et al. \cite{kainz2015fast} could be applied prior to brain segmentation.
 
In this work, segmentation was performed using a multi-scale CNN as proposed by Moeskops et al. \cite{moeskops2015automatic}. Similar multi-scale CNNs could likely also be used, such as the network proposed by Kamnitsas et al. \cite{kamnitsas2017efficient}. In addition, other segmentation network architectures could be evaluated, such as the fully convolutional architectures as proposed by Shelhamer et al. \cite{shelhamer2017fully} and Ronneberger et al. \cite{ronneberger2015u}, which contain upsampling layers and skip-connections to acquire multi-scale information. Another approach used for segmentation are dilated CNNs \cite{Yu16,wolterink2016dilated}, which can achieve large receptive fields with a limited number of trainable weights. Moreover, in future work, network architectures that use 3D information will be investigated.

The gestational age of fetuses included in this study was relatively narrow (range: $22.9$ to $34.6$). However, the method can likely be applicable to fetal MR scans made at other gestational ages. Our future work will investigate application of this method in a large set of fetal scans acquired in a broad range of gestational ages.
 
\section{Acknowledgements}{This study was sponsored by the Research Program Specialized Nutrition of the Utrecht Center for Food and Health, through a subsidy from the Dutch Ministry of Economic Affairs, the Utrecht Province and the Municipality of Utrecht.}
	
	%
	%
	
	\bibliographystyle{splncs03}
	\renewcommand{\bibname}{\protect\leftline{Bibliography}}

	\bibliography{semisupervised}

\end{document}